\documentclass{llncs}
\usepackage{amsmath}
\usepackage[T1]{fontenc}
\usepackage{graphicx}
\usepackage{booktabs}
\begin{document}
\title{RIRO: Reshaping Inputs, Refining Outputs Unlocking the Potential of Large Language Models in Data-Scarce Contexts}
\author{
    Ali Hamdi\inst{1,2} \and 
    Hozaifa Kassab\inst{1} \and 
    Mohamed Bahaa\inst{1} \and 
    Marwa Mohamed\inst{2}
}
\institute{
    AiTech AU, Preston, Australia\\
    \email{ali,hothaifa,mbahaa@aitech.net.au} \and
    Faculty of Computer Science, MSA University, Egypt\\
    \email{ahamdi,mmsolayman@msa.edu.eg}
}

\maketitle

\begin{abstract}
Large language models (LLMs) have significantly advanced natural language processing, excelling in areas like text generation, summarization, and question-answering. Despite their capabilities, these models face challenges when fine-tuned on small, domain-specific datasets, often struggling to generalize and deliver accurate results with unfamiliar inputs. To tackle this issue, we introduce RIRO, a novel two-layer architecture designed to improve performance in data-scarce environments. The first layer leverages advanced prompt engineering to reformulate inputs, ensuring better alignment with training data, while the second layer focuses on refining outputs to minimize inconsistencies. Through fine-tuning models like Phi-2, Falcon 7B, and Falcon 1B,  with Phi-2 outperforming the others. Additionally, we introduce a benchmark using evaluation metrics such as cosine similarity, Levenshtein distance, BLEU score, ROUGE-1, ROUGE-2, and ROUGE-L. While these advancements improve performance, challenges like computational demands and overfitting persist, limiting the potential of LLMs in data-scarce, high-stakes environments such as healthcare, legal documentation, and software testing.  
\end{abstract}

\section{Introduction}

Generating consistent and accurate outputs from large language models (LLMs) fine-tuned on small datasets introduce a significant challenge, leading to the exploration of wide range of solutions aims to improve their generalization abilities \cite{raiaan2024review}. One common method involves data augmentation techniques, such as paraphrasing and back translation, to artificially expand the training dataset \cite{pellicer2023data}. While these strategies can increase data diversity, they come with common drawbacks. Paraphrasing often introduces noise, and adds imperfect or ambiguous examples into the dataset, which negatively impacts the model performance \cite{di2023investigating}. Similarly, back-translation, though effective for certain tasks, risks missing the meaning of the original inputs, further complicating the model’s ability to produce accurate and reliable outputs \cite{mohamed2024impact}.

\begin{figure}[ht]
    \centering
    \includegraphics[width=\linewidth]{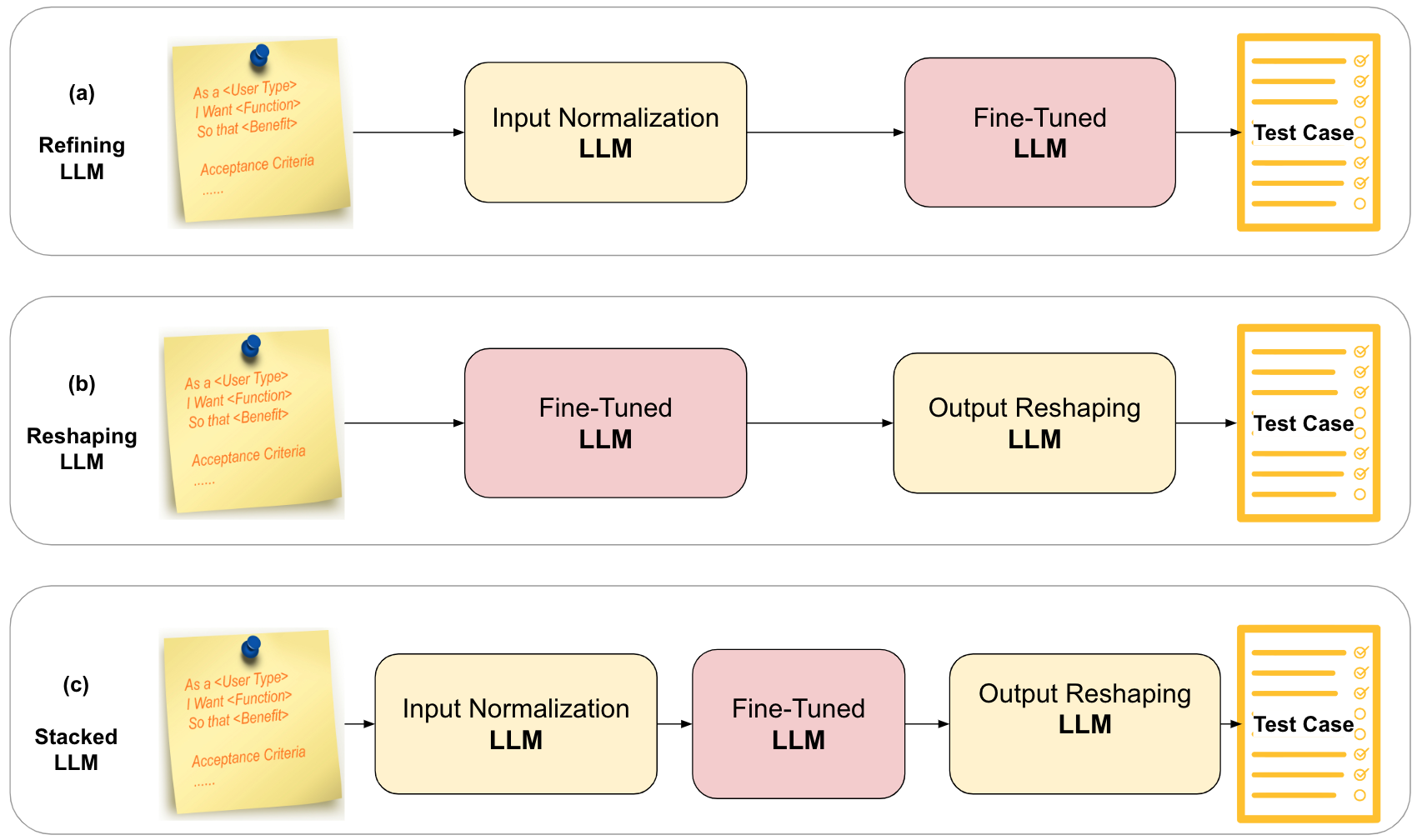}
    \caption{The proposed model architectures for RIRO Versions. 
    (a) Refining LLM: This architecture focuses on input normalization. It aligns the input user stories with the training data distribution. 
    (b) Reshaping LLM: Here, the output reshaping layer to ensure coherent test cases. This method adjusts the final output to maintain consistency and accuracy. 
    (c) Stacked LLM: A combined approach that first normalizes the input, passes it through a fine-tuned LLM, and applies reshaping for output generation.}
    \label{fig:LLM}
\end{figure}

A more structured alternative involves enforcing fixed input formats using predefined templates or rule-based systems \cite{mcroy2003augmented}. Although this can ensure consistency in inputs, it sacrifices flexibility and scalability, requiring users to manually adapt inputs to fit to strict guidelines \cite{xue2024repeat,kleppmann2017designing}. This approach is impractical for real-world applications where input variability is unavoidable, limiting its effectiveness across diverse contexts \cite{saha2024llm}.

Despite these efforts, producing high-quality, structured outputs from LLMs trained on small datasets remains an unresolved issue \cite{raiaan2024review}. Current solutions either adds unwanted noise,  which demand extensive computational resources, or restrict constraints on inputs, making them unsuitable for many real-world applications \cite{fioretto2018distributed}. There is a need for  more efficient and robust approach that can handle input variability while preserving output quality from fine-tuned LLMs in data-scarce environments \cite{yuan2023revisiting}.

To address these challenges, we propose a novel two-step approach that leverages the strengths of LLMs while maintaining efficiency and adaptability. RIRO employs two LLM layers working in Adjective way. The first LLM reformulates the input, ensuring that it aligns with the structure and format expected by the model during fine-tuning, irrespective of how the input is initially phrased. The second LLM, fine-tuned on a small, domain-specific dataset using Quantized Low-Rank Adaptation (QLoRA), processes the reformulated input to generate accurate and consistent outputs. By standardizing the input through reformulation, we minimize variability, enabling the fine-tuned model to focus on producing high-quality outputs without inconsistencies.

As illustrated in Figure \ref{fig:LLM}, RIRO is composed of three model architectures. The first architecture, Refining LLM, focuses on input normalization followed by fine-tuning to generate the final test cases. The second architecture, Reshaping LLM, introduces an output reshaping layer after the fine-tuning process to maintain coherence in the generated test cases. The final architecture, Stacked LLM, combines both input normalization and output reshaping for enhanced performance in data-scarce environments. Fine-tuning was performed using several models, while Phi-2 as a backbone LLM demonstrating superior performance. 

RIRO effectively addresses the limitations of prior methods by focusing on the input level. This enables improved generalization, greater flexibility, and more reliable performance from the fine-tuned LLM, even with limited training data. Furthermore, the use of QLoRA allows for efficient model fine-tuning, significantly reducing the computational costs typically associated with such processes. The result is a scalable and practical solution that can be applied across various domains where data scarcity is common, but accuracy and precision remain critical.

\section{Related Works}
The automation of software testing has long been an area of interest in both academia and industry, aimed at ensuring that applications function correctly and meet user expectations \cite{garousi2020software,tyagi2020intelligent}. One prevalent approach in software testing is designing test cases based on user stories \cite{carroll2003making}. User stories are concise, natural language descriptions of features from the end-user perspective, outlining the desired functionalities of the software \cite{kamalakar2013automatically}. Transforming these user stories into test cases involves interpreting user requirements and creating executable scenarios that validate the software's behavior \cite{barnum2020usability}. This process helps bridge the gap between user expectations and technical implementation, thereby enhancing software quality and reliability \cite{hertzum2022usability}.

However, manually generating test cases from user stories is a time-consuming and error-prone task \cite{klammer2017journey}. Traditional methods heavily depend on the tester's expertise to interpret and convert user narratives into formal test cases, which often leads to inconsistencies and omissions \cite{stray2022exploring}. To address these issues, research has focused on automating this process using natural language processing (NLP) and machine learning techniques \cite{chinnaswamy2024user}. Early approaches predominantly used rule-based systems and pattern matching to extract test cases from requirements \cite{ichii2012rule}. Although these methods introduced some level of automation, they struggled with the variability and ambiguity that are inherent in natural language \cite{bhatt2021case}.

NLP models have been utilized to advance a wide-range of tasks \cite{hamad2022attention,badaro2019survey,baly2017omam,hamdi2018clasenti,hamdi2016review,hamad2022steducov,hamad2024asem,hamad2022empathy}. The advent of Large Language Models (LLMs), such as GPT-based architectures, has introduced new possibilities in automating test case generation \cite{karmarkar2024navigating}. LLMs, trained on vast datasets, have demonstrated significant success in understanding and generating human-like text \cite{wang2022towards}. These models have been applied to a range of natural language tasks, including the extraction of test cases from user stories \cite{chuor2024user}. Recent works have shown that LLMs can effectively manage more complex and nuanced user stories, reducing the reliance on manual generation efforts \cite{schafer2023empirical}. For instance, LLMs have been employed to automate various testing processes such as unit tests, integration tests, and acceptance tests by transforming user stories directly into executable test scripts \cite{alshahwan2024automated}.

Despite their potential, challenges remain in deploying LLMs for this task \cite{xue2024domain}. LLMs often produce outputs that are too general or misaligned with the specific requirements of the software being tested \cite{ryan2024code}. Additionally, single-pass LLM approaches, which generate test cases in one run, tend to overlook important edge cases or fail to account for complex relationships within user stories \cite{hurani2024investigating}. This has led to an exploration of advanced techniques aimed at improving the quality and accuracy of LLM-generated test cases \cite{gu2024testart}.

Our contribution introduces a novel approach using stacked layers of LLMs to reshape inputs and refine outputs, significantly enhancing test case generation. Unlike traditional single-pass methods, RIRO processes user stories through multiple LLM layers, with each layer focused on either input standardization or output refinement. This multi-layered architecture enables the model to better handle the variability of natural language and produce more accurate and comprehensive test cases. The additional layers iteratively refine the generated outputs, leading to more precise and actionable test cases that align with the functional and technical requirements of the software. This approach demonstrates superior performance over existing state-of-the-art methods, showcasing the potential of stacked LLM architectures in automating complex software testing tasks.

\section{Dataset}
The dataset used in this study is a subset of the user story neodataset, which consists of issues reported in a software development project. Each issue in the dataset includes fields such as the title, description, and story points. This rich collection of issue data provides a robust foundation for analyzing project management practices, assessing workload estimations, and evaluating the progress and quality of the software system. By examining these issues, we can gain insights into the effectiveness of issue tracking and resolution processes in ensuring the proper functionality and overall success of the project.

\section{Methodology}

In this section, we present a novel approach for generating high-quality outputs from large language models (LLMs) fine-tuned on small, domain-specific datasets. RIRO leverages a layered framework combining reformulation, fine-tuning and output reshaping. The approach is applied to the Phi-2 LLM for experimentation, demonstrating its generalizability and effectiveness across diverse language tasks.

\subsection{Phi-2 LLM Architecture}
RIRO  is designed to be applicable to various LLM architectures, though for our experiments, we utilize the Phi-2 LLM, an enhanced version of GPT-based models. The Phi-2 LLM incorporates several improvements for handling complex and diverse language tasks, especially when the model is fine-tuned on small datasets. 

The core architecture is based on the transformer model, characterized by its multi-head self-attention mechanism \cite{vaswani2017attention}. The attention mechanism aggregates information from different positions in the input sequence:
\[
\text{Attention}(Q, K, V) = \text{softmax}\left(\frac{QK^T}{\sqrt{d_k}}\right)V
\]
where \( Q \), \( K \), and \( V \) represent the query, key, and value matrices, and \( d_k \) is the dimensionality of the keys. Phi-2 enhances the standard transformer by employing adaptive positional encodings, which dynamically adjust based on the complexity and length of the input sequences.

\paragraph{Adaptation for Small Datasets}
The objective is to adapt the model without overfitting, which is a common issue with smaller datasets. QLoRA allows for fine-tuning a subset of the model's parameters, denoted by \( \Delta \Theta \), while preserving computational efficiency.

The total parameter space of the model \( \Theta \) is reduced during fine-tuning, with a rank-reduction technique applied to minimize the number of updated parameters:
\[
N_{QLoRA} = r \times k, \quad r \ll N
\]
where \( r \) represents the rank of the low-rank approximation, and \( k \) is the number of fine-tuned parameters.

During fine-tuning, the pre-trained model's weights \( W \) are quantized to a lower precision (e.g., 4-bit):
\[
\hat{W} = W_q + UV^T
\]
where \( W_q \) are the quantized weights, and \( UV^T \) represents the low-rank adaptation applied to fine-tune the model with minimal resource overhead.

\subsection{LLM-based Reformulation}
The reformulation step ensures that the input is normalized before passing through the model. Given a raw input query \( x \), the reformulation process standardizes the input into a format \( x' \) that matches the structure seen during fine-tuning:
\[
x' = f_r(x), \quad f_r: X \to X'
\]
where \( X \) is the space of raw inputs, and \( X' \) is the space of normalized inputs. This normalization mitigates variability in phrasing and structure, ensuring that the model can more effectively handle the input.

For example, if the user story input follows a format of "Action, Condition, Result," the reformulation layer restructures any raw user story to align with this template, improving the model's ability to generate accurate test cases.

\subsection{QLoRA Fine-Tuning}
The reformulated input \( x' \) is then passed to the fine-tuned LLM. Fine-tuning is performed using QLoRA \cite{dettmers2023qlora}, which is specifically designed for efficient fine-tuning on small datasets while maintaining model performance.

The model's parameters \( \theta \) are first quantized:
\[
\theta_q = \text{Quantize}(\theta)
\]
Low-rank adaptation further updates only a subset of parameters, \( \Delta \theta \), relevant to the task, leading to:
\[
\theta' = \theta_q + \Delta \theta
\]
This approach allows the model to adapt to domain-specific requirements without the computational overhead typically associated with fine-tuning large LLMs.

\subsection{LLM-based Reshaping}
The final step in the process involves reshaping the generated output to enhance readability and alignment with the required format. Let \( y \) represent the initial output generated by the fine-tuned LLM, and \( y' \) represent the reshaped output:
\[
y' = \text{Reshape}(y)
\]
This step ensures that the generated test cases are not only accurate but also structured in a manner that makes them easier to interpret and integrate into the software testing process.

\subsection{Ablation Study: Experimental Variants}
To validate RIRO, we conduct an ablation study that evaluates three different variants of the approach:
\begin{itemize}
    \item \textbf{LLM-RFR (Reformulation-Fine-tuning-Reshaping):} This variant represents the full pipeline, including reformulation, fine-tuning, and reshaping.
    \item \textbf{LLM-RF (Reformulation-Fine-tuning):} This variant excludes reshaping to isolate the effects of reformulation and fine-tuning.
    \item \textbf{LLM-FR (Fine-tuning-Reshaping):} This variant omits reformulation, evaluating the importance of normalizing the input before fine-tuning.
\end{itemize}

The ablation study reveals that the full pipeline (LLM-RFR) consistently outperforms the other variants, particularly in terms of accuracy and consistency of the generated test cases. The inclusion of reformulation and reshaping significantly enhances the model’s ability to handle diverse inputs.

\subsection{Evaluation}
We evaluate RIRO using the following metrics:
\begin{itemize}
    \item \textbf{BLEU Score:} Measures the overlap of \( n \)-grams between the generated output and reference text.
    \item \textbf{ROUGE-1, ROUGE-2, ROUGE-L:} Assess unigram, bigram, and longest common subsequence overlap, respectively.
    \item \textbf{Levenshtein Distance:} Quantifies the minimum number of edits required to transform one string into another.
    \item \textbf{Cosine Similarity:} Measures the cosine of the angle between two non-zero vectors, typically used to assess the similarity between text sequences.
\end{itemize}

These metrics ensure a comprehensive evaluation of both the syntactic and semantic accuracy of the generated outputs, demonstrating the effectiveness of RIRO in automating the generation of test cases from user stories.

\textbf{In conclusion}, RIRO combines the strengths of LLM-based reformulation, QLoRA fine-tuning, and reshaping to create a robust framework for automating test case generation. By employing a layered approach, we enhance the precision, consistency, and relevance of the outputs, advancing the state of the art in automated software testing.

\section{Results and Discussion}

Table 1 presents a comparative performance analysis of the three proposed architectures: Reshaping (input-focused), Refining (output-focused), and RIRO (combining both) across various evaluation metrics, including BLEU, ROUGE (F1), Levenshtein Distance, and Cosine Similarity. 
\begin{table}[h]
\centering
\caption{Performance comparison across various evaluation metrics.}
\begin{tabular}{lccccc}
\toprule
\textbf{Metric} & \textbf{Phi-2} & \textbf{ Reshaping} & \textbf{ Refining} & \textbf{ 
 RIRO} \\
\midrule
BLEU Score & 0.55 & 0.66  & 0.62 & 0.72 \\
ROUGE1 (F1) & 0.265 & 0.310  & 0.375 & 0.402 \\
ROUGE2 (F1) & 0.128 & 0.122  & 0.147 & 0.149 \\
ROUGE L (F1) & 0.172 & 0.202  & 0.227 & 0.257 \\
Levenshtein Distance & 1157.620 & 1157.080  & 1420.500 & 1000.880 \\
Cosine Similarity & 0.816 & 0.826  & 0.849 & 0.891 \\
\bottomrule
\end{tabular}
\end{table}
The results highlight the strengths and limitations of each approach, with RIRO emerging as the most robust model.

\textbf{BLEU Score:} The RIRO model achieves the highest BLEU score (0.72), surpassing both Reshaping (0.66) and Refining (0.62). This indicates that RIRO produces text with the greatest n-gram overlap, meaning it excels in syntactic and lexical accuracy. The combination of input reshaping and output refining allows RIRO to capture both the surface structure and finer nuances of the text, leading to higher precision in matching the reference. 

\textbf{ROUGE Scores:} The RIRO model consistently outperforms the other two architectures in all ROUGE metrics, underscoring its superior ability to capture overlapping content between the generated text and the reference. 
\textbf{ROUGE-1 (F1)}, RIRO (0.402) surpasses Reshaping (0.310) and Refining (0.375), showing that it generates outputs with better word-level recall. \textbf{ROUGE-2 (F1)}, while both Reshaping (0.122) and Refining (0.147) offer competitive performance, RIRO again leads with 0.149, reflecting its ability to better maintain coherence across bi-grams and multi-word phrases. 
\textbf{ROUGE-L (F1)}, RIRO (0.257) demonstrates superior long-sequence alignment, outperforming Reshaping (0.202) and Refining (0.227). This highlights RIRO’s capacity for capturing longer, syntactically cohesive patterns. 

\textbf{Levenshtein Distance:} RIRO also exhibits the lowest Levenshtein Distance (1000.880), compared to Reshaping (1157.080) and Refining (1420.500). A lower Levenshtein score indicates fewer character-level edits are required to transform RIRO’s output into the reference text, making it the most accurate in terms of exact character matching. This suggests that while Reshaping improves character-level precision, the combination of both input reshaping and output refining in RIRO leads to even greater accuracy. 

\textbf{Cosine Similarity:} The Cosine Similarity score for RIRO (0.891) is the highest, exceeding that of Reshaping (0.826) and Refining (0.849). This score reflects how well the generated text captures the semantic meaning of the reference. RIRO's higher score indicates that the combination of reshaping and refining enables it to retain both the overall content and meaning of the text more effectively than either approach alone. 

\textbf{Discussion}:

The results demonstrate that the \textbf{Reshaping} architecture is effective at enhancing the syntactic structure of the generated text, as reflected in its relatively high BLEU and ROUGE-1 scores. However, it lacks in semantic coherence, as shown by its lower ROUGE-2 and Cosine Similarity scores.

The \textbf{Refining} architecture, focused on refining outputs, performs better than Reshaping in capturing longer sequences and semantic content, as seen in its improved ROUGE-2, ROUGE-L, and Cosine Similarity scores. However, it does not achieve the same level of lexical accuracy as Reshaping, as evidenced by its slightly lower BLEU score.

The \textbf{RIRO model}, which combines both reshaping inputs and refining outputs, demonstrates superior performance across all metrics. The combination of both techniques allows RIRO to balance lexical and semantic quality, leading to a more robust output that captures both structural precision and meaningful content. This makes RIRO the most effective model, as it consistently surpasses the other architectures in both surface-level and deeper semantic evaluations.

In conclusion, while both Reshaping and Refining offer distinct benefits, the RIRO model proves to be the most comprehensive and effective approach, offering significant improvements in generating text that is both syntactically accurate and semantically coherent.

\section{Conclusion}
In this work, we introduced the Stacked-LLM architecture, which significantly improves upon existing text generation methods by addressing key limitations in fine-tuning large language models on small, domain-specific datasets. Through a combination of input normalization, fine-tuning, and output reshaping, our approach achieves superior performance across critical evaluation metrics, such as BLEU, ROUGE, and cosine similarity. These results highlight the practical benefits of our architecture, particularly in data-scarce environments where maintaining consistency and accuracy is crucial. Moreover, this work offers valuable theoretical contributions to the field of natural language processing, providing a scalable and flexible solution for enhancing language model performance in specialized applications. Our findings represent a meaningful advancement in both the practical application and theoretical understanding of LLMs, setting the stage for further exploration and optimization in this rapidly evolving domain.

\bibliographystyle{bibtex/spmpsci}
\bibliography{ref}
\end{document}